\documentclass[10pt,conference]{IEEEtran}
\IEEEoverridecommandlockouts
\usepackage{cite}
\usepackage{soul}
\usepackage{amsmath,amssymb,amsfonts}
\usepackage{algorithmic}
\usepackage{graphicx}
\usepackage{textcomp}
\usepackage{xcolor}
\def\BibTeX{{\rm B\kern-.05em{\sc i\kern-.025em b}\kern-.08em
    T\kern-.1667em\lower.7ex\hbox{E}\kern-.125emX}}
\begin{document}

\title{Fair Resource Allocation for Fleet Intelligence\\

\thanks{\textcolor{black}{This work was supported in part by the National Science Foundation grants under No. 2148186 and No. 2133481. Any opinions, findings, and conclusions or recommendations expressed in this material are those of the authors and do not necessarily reflect the views of the National Science Foundation}.}
}

\author{\IEEEauthorblockN{Oguzhan Baser, Kaan Kale, Po-han Li, and Sandeep Chinchali}
\IEEEauthorblockA{Wireless Networking Communications Group (WNCG)\\
Department of Electrical and Computer Engineering\\
The University of Texas at Austin\\
Email: {\tt oguzhanbaser@utexas.edu}
}}

\maketitle

\begin{abstract}
Resource allocation is crucial for the performance optimization of cloud-assisted multi-agent intelligence. Traditional methods often overlook agents' diverse computational capabilities and complex operating environments, leading to inefficient and unfair resource distribution. To address this, we open-sourced Fair-Synergy, an algorithmic framework that utilizes the concave relationship between the agents' accuracy and the system resources to ensure fair resource allocation across fleet intelligence. We extend traditional allocation approaches to encompass a multidimensional machine learning utility landscape defined by model parameters, training data volume, and task complexity. We evaluate Fair-Synergy with advanced vision and language models such as BERT, VGG16, MobileNet, and ResNets on datasets including MNIST, CIFAR-10, CIFAR-100, BDD, and GLUE. We demonstrate that Fair-Synergy outperforms standard benchmarks by up to 25\% in multi-agent inference and 11\% in multi-agent learning settings. Also, we explore how the level of fairness affects the least advantaged, most advantaged, and average agents, providing insights for equitable fleet intelligence.
\end{abstract}

\section{Introduction}
Consider a scenario in which a fleet of agents operates in heterogeneous environments, each with varying on-device memory, compute power, computation time, or a combination thereof (henceforth referred to as resources). These agents have their local machine learning (ML) models with accuracy limited by their local resources. Hence, they can collaborate with a cloud, a central server offering additional resources, to boost their accuracy. Ideally, agents should offload all their tasks to the cloud to achieve the best possible performance. However, the cloud has limited capacity, so accommodating all these requests results in latency and unfair resource allocation. Fairness in this context means that no single agent can monopolize the cloud, leading to a balanced allocation where all agents improve their performance without significantly disadvantaging any single agent. Our key question of interest is ``\textit{How do we fairly allocate the cloud resources among a fleet of agents to maximize the aggregate accuracy?}'' We answer this question in the context of both Distributed Learning (DL) and Real-Time Inference (RTI), where agents must continuously adapt to changing environments and coordinate in latency-sensitive tasks that demand timely and efficient resource allocation.

\begin{figure}[t]
    \centering
    \includegraphics[width=0.9\linewidth]{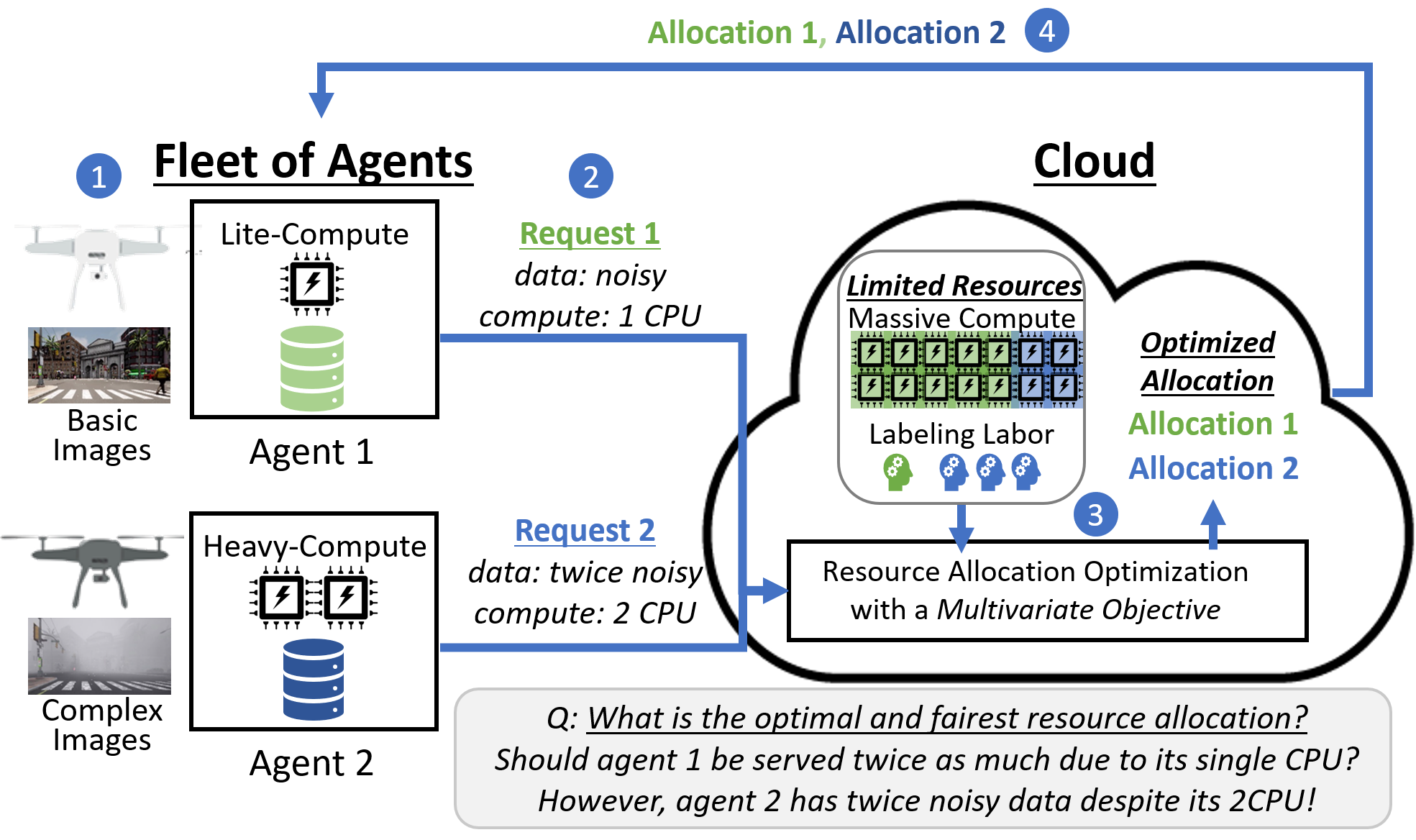}\caption{ \small\textbf{\textit{Nuanced Resource Allocation for Fairness:}} Agent 1 (1 CPU) processes less noisy data, while agent 2 (2 CPUs) handles data that is twice as noisy (1). Both agents request cloud resources to enhance performance (2). In response, the cloud optimizes allocation based on data complexity and local computes (3), assigning resources to each agent accordingly (4). Intuitively, while agent 1 might seem entitled to twice the resources due to having half the compute power of agent 2, the complex data handled by agent 2 makes fair allocation nontrivial. Increasing the resource dimension, such as the amount of labeled data each agent locally has and how the cloud distributes annotations, makes it even more nuanced.\vspace{-3mm}}
    \label{fig:toyex}
\end{figure}

Today's resource allocation mechanisms for DL and RTI systems often lead to suboptimal allocations because they do not consider the differences in agents' local resources and the heterogeneity of the environment in which they operate. Consequently, this leads to delayed, costly, and inequitable services for these agents. Consider the motivating example in Fig. \ref{fig:toyex}. Assume there are two camera-equipped drones and a shared cloud. Given that drone 1 has lite-compute (less memory and less accurate models) and drone 2 has heavy-compute (double memory for bigger and more accurate models), the fairness criterion dictates that the cloud should serve more to drone 1 than drone 2. However, the situation complicates if drone 2 captures harder-to-parse images compared to drone 1. In this case, a fair allocation must account for both the computational load imposed by the models' sizes and the captured images' inherent difficulty by balancing these factors.

We conduct extensive experiments on state-of-the-art computer vision and language datasets to empirically show \textbf{a concave relationship between accuracy and system resources}, such as the model size (scaling with the number of parameters), the dataset size (scaling with augmentation), or the processing time (scaling with model depth). This concave relationship naturally makes sense due to overfitting for increasing model complexity and reduced marginal information gain from larger datasets.

We combine this insight with Network Utility Maximization (NUM), a well-studied framework \cite{kelly1998rate} in networking. \textcolor{black}{It allocates limited channel capacity among data flows to maximize users' overall satisfaction.} Each user has a unique utility function that reflects satisfaction with the allocated flow. For example, streaming HD video requires more bandwidth than file transfers. Our key observation is that similar resource allocation challenges exist in intelligent fleets. They must share the cloud resources, and their overall task accuracy experiences diminishing marginal returns as the cloud allocates more resources (e.g., compute power). To the best of our knowledge, we are the first to draw this analogy between NUM and resource allocation in fleet intelligence based on their concave objectives. NUM provides a structured approach for fair resource distribution, preventing monopolization, and enhancing collective performance. However, unlike the simple convex utility functions typical in NUM, multi-agent ML often involves complex utility landscapes, which we address and extend in this paper.



We introduce {\tt Fair-Synergy}, a general algorithmic framework to determine the optimal and fair resource allocation in multi-agent DL and RTI. In light of prior work, our contributions are fourfold. 
\begin{itemize}
\item We \textbf{empirically show the concave relationship} between an agent's ML performance and its resources. Through extensive testing with advanced Computer Vision (CV) and Language Models (LMs) on real-world datasets, \textbf{we affirm the potential of a NUM-like framework} as an effective allocation method.
\item We model ML task complexity by correlating accuracy with model size (parameters) and data diversity (volume/augmentation). Then, we craft \textbf{a multivariate utility function} to guide \textbf{our theoretical resource allocation framework} within multi-agent systems.
    \item We \textbf{mathematically derived the first fairness condition for cloud-assisted multi-agent ML settings}. It dictates a precise resource allocation for each agent, considering agents' resources and task complexities.
    \item We present extensive results demonstrating the performance of our open-sourced\footnote{{\tt \scriptsize \textcolor{black}{https://github.com/UTAustin-SwarmLab/Fair-Synergy}}\label{coderef}} tool, Fair-Synergy, in resource-shared fleet intelligence. The results reveal the \textbf{fairness-performance trade-off} and we surpass the benchmarks by up to \textbf{25\%} in RTI and \textbf{11\%} in DL, with even greater margins as the system scales.
\end{itemize}




\section{Background}
Before the problem formulation, we outline \textcolor{black}{the} key concepts and insights shaping our multi-agent ML utility function.

\begin{figure}[!t]
    \centering
    \includegraphics[width=\linewidth]{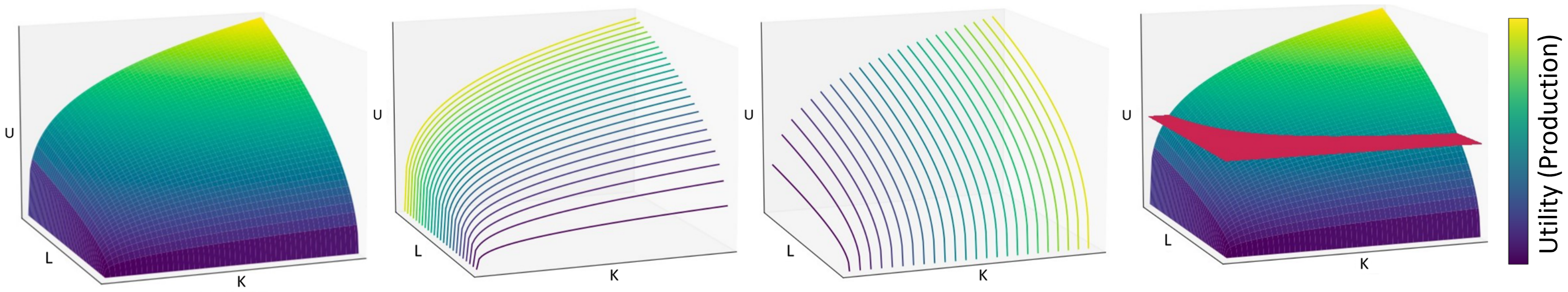}\caption{\small
    \textbf{Multivariate Utility:} We visualize the Cobb-Douglas function in Eq. \ref{eqn:multidimutility} (left). It exhibits diminishing marginal returns for increasing resources. This concave behavior is more obvious when one of the variables is kept constant (middle). The red plane (right) depicts iso-production levels for the variables such as capital and labor. It shows that the same level of utility (e.g., production) can be obtained by trading off the resources. This serves as a foundation for the formulation of our multivariate optimization objective.\vspace{-3mm}}
    \label{fig:cobbdouglas}
\end{figure}

\begin{figure*}[t]
    \centering \includegraphics[width=0.39\linewidth]{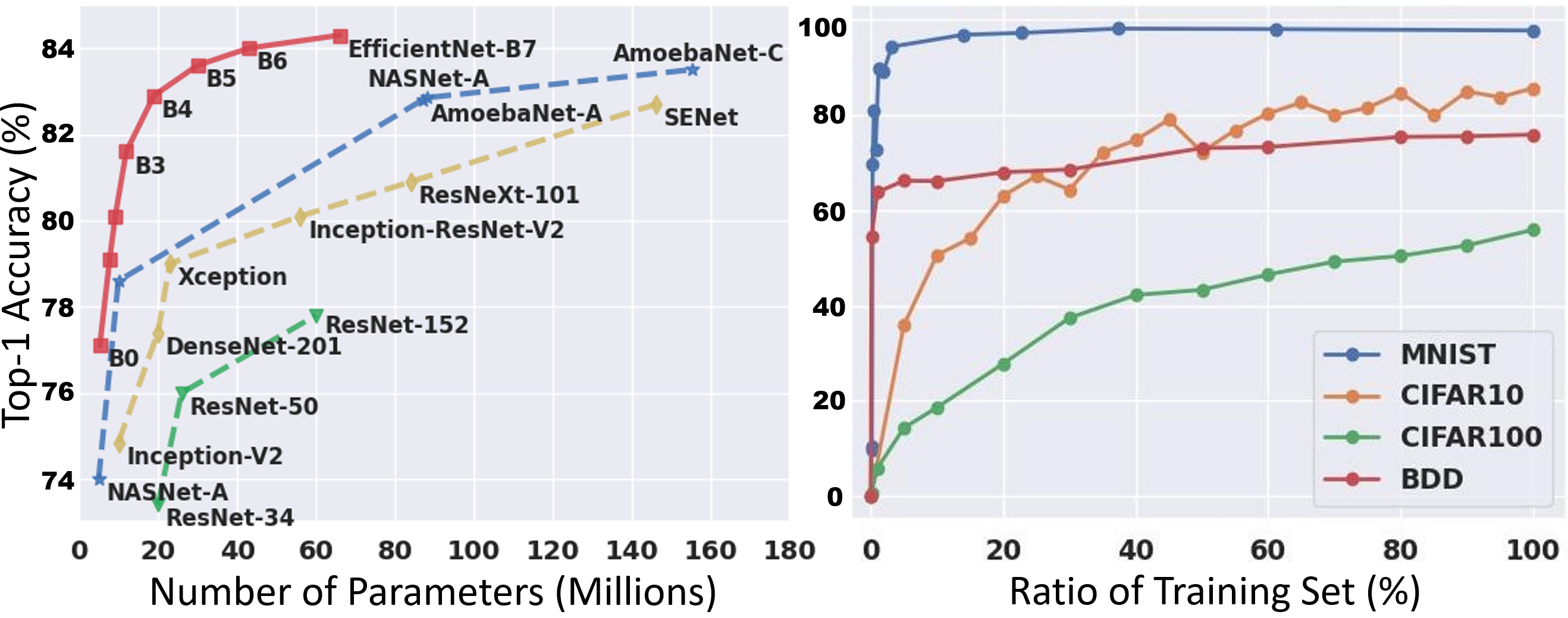}\includegraphics[width=0.61\linewidth]{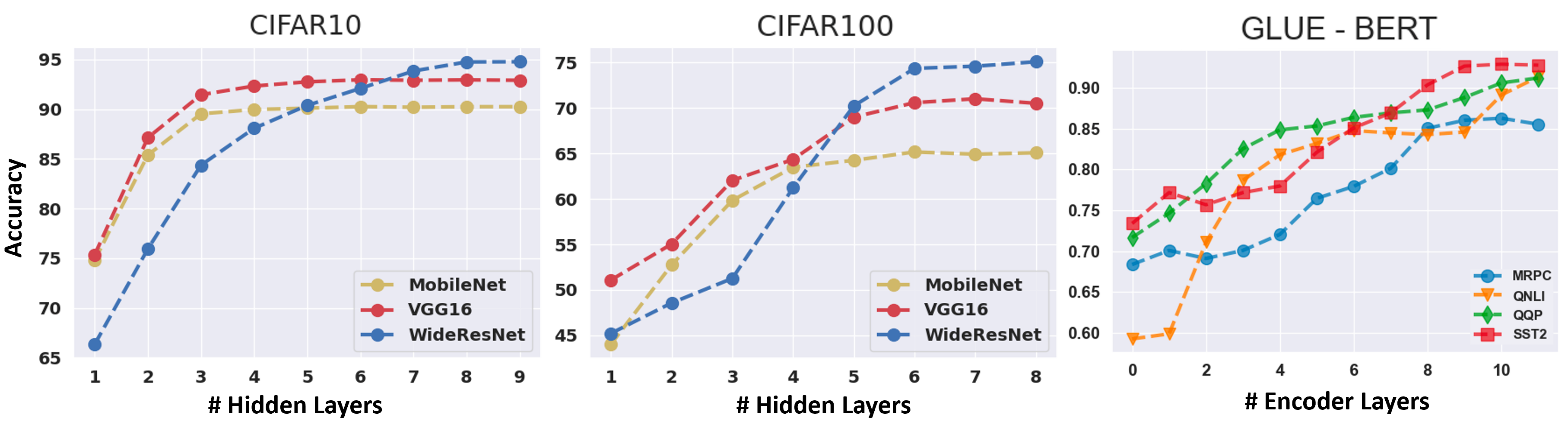}\caption{\small \textbf{Concave Model Accuracy vs. ML Resources} \textit{In training (left two)}, increasing model size (EfficientNet \cite{tan2019efficientnet} on ImageNet) or training data volume (ResNet-32 on MNIST, CIFAR-10, CIFAR-100, BDD \cite{yu2020bdd100k}) yields diminishing accuracy gains, demonstrating concavity. \textit{In testing (right three)}, our observations on the {\tt CIFAR-10} (left), {\tt CIFAR-100} (middle), and {\tt GLUE} (right) datasets reveal a concave trend in model performance for the increasing compute time (number of hidden layers). This pattern holds across diverse model architectures ({\tt MobileNet}, {\tt VGG16}, {\tt WideResNet} in CV, and {\tt BERT} in LM) and language tasks such as {\tt MRPC, QNLI, QQP,} and {\tt SST2} (right). {\tt MRPC} evaluates semantic equivalence, {\tt QNLI} tests textual entailment, {\tt QQP} identifies duplicate questions, and {\tt SST2} analyzes sentiments.}
    \label{fig:concvres}\vspace{-3mm}
\end{figure*}

\underline{\textit{Concave Accuracy as a Function of agents' ML Resources}:}\\
We posit that one can perceive an agent's ML accuracy as its utility. The utility function is naturally concave due to the law of diminishing returns; as agents are allocated more resources (e.g., model size, computation power, training data), their performance improves, but the rate of improvement decreases. Consider agents in an Amazon warehouse tasked with sorting and packing items. Increasing \textcolor{black}{the} compute power (e.g., adding more CPUs/GPUs) allows bigger ML models, and thus, higher accuracy for object recognition and path planning. However, further increasing it yields only marginal gains beyond a point due to limiting factors, such as sensor resolution, becoming bottlenecks. We empirically show the evidence supporting our intuition in Fig. \ref{fig:concvres}. Experiments with various ML architectures across CV and LM tasks consistently show a concave resource-performance relationship due to diminishing marginal returns for increasing resources.

\noindent\textit{\underline{Cobb-Douglas Production Function as a Multivariate Utility}:} After demonstrating that accuracy is a concave function of the agents' ML resources, our next step is to form a utility function that estimates the accuracy as a multivariate function of these resources. We introduce a \textbf{\textit{multivariate utility function}} tailored for multi-agent ML systems, drawing from the Cobb-Douglas production function \cite{varian2003intermediate} from microeconomics.

As an intuitive example, imagine a factory where the overall objective is production, and it depends on two resources: capital and labor. If the amount of labor remains fixed, increasing capital will initially boost production, but the rate of increase will eventually diminish. The same occurs for increasing labor while keeping capital fixed. Analogously, in ML, the utility is accuracy, and the different resources are model size, scaling with computational power, and the amount of training data. Increasing training data size for fixed compute power will initially enhance accuracy, but the improvements will eventually taper off. This makes the Cobb-Douglas production function \textit{\textbf{a natural fit for ML}}. We are the first to draw this analogy, highlighting its suitability for estimating ML performance in multi-agent systems. We illustrate the Cobb-Douglas function in Fig. \ref{fig:cobbdouglas}. It is defined as follows.


\textit{Definition 1: Given an investment of capital $K$ and applied labor $L$, the total production, denoted as $U$, is defined in the Cobb-Douglas form as:}
\begin{equation}\small
     U(K,L;\gamma_K,\gamma_L) = K^{\gamma_K}*L^{\gamma_L};\quad \forall i, \gamma_i \in (0,1).
     \label{eqn:multidimutility}
\end{equation}
\textit{where $\gamma_K$ and $\gamma_L$ represent capital and labor elasticity.}

$\gamma$ quantifies the impact of resource on production. Smaller $\gamma$ indicates sharper diminishing returns. $\gamma_K > \gamma_L$ prioritizes capital (e.g., tech), while $\gamma_L > \gamma_K$ emphasizes labor (e.g., farming). $\gamma_K + \gamma_L = 1$ reflects constant returns (doubling inputs doubles output), and $\gamma_K + \gamma_L > 1$ indicates increasing returns (doubling inputs more than doubles output).


\noindent\textit{Remark 1: Provided that the variables remain independent, the Cobb-Douglas utility function $U$ can extend to multiple variables, each with its own distinct elasticity $\gamma$, shown as:}
\begin{equation}\small
     U(K_1, ..., K_n; \gamma_1, ... , \gamma_n) = K_1^{\gamma_{1}}*K_2^{\gamma_{2}}*...*K_n^{\gamma_{n}};\forall i, \gamma_i \in (0,1).
     \label{eqn:multidimutilityextended}
\end{equation}
This function accommodates all independent resource types in a single agentic metric: accuracy. This way, we can capture the intricate relationship between ML hyperparameters in multi-agent ML systems. The remark states that the number of independent variables in our utility can flexibly expand to approximate the model performance and its correlation with the available types of resources.


\textit{\underline{Network Utility Maximization (NUM):}}
NUM is a framework in networking for fair and efficient resource allocation. It addresses the distribution of limited channel capacity (e.g., bandwidth) among multiple users \cite{nagaraj2016numfabric}. 
It allocates the capacity $C$ as data flows $x_i$ among multiple users. Each user has a utility function $U_i$ that represents their satisfaction based on the received bandwidth.
The utility is typically concave due to diminishing returns with increasing resources. The goal is to maximize the total utility across users. It is formulated as:
\begin{equation}
\begin{aligned}
    \min_{x_{i}}&\sum_{i=1}^{n} -U_{i}(x_{i}) &&\text{\scriptsize(Maximize Total Utility)} \\
    \textrm{s.t.}&(\sum_{i=1}^{n} x_{i}) - C \leq 0,&\forall i\in\{1,...,n\},& \quad\hspace{2.5mm} \text{\scriptsize(Limited Channel)}\\
    &-x_{i} \leq 0,& \forall i\in\{1,...,n\}. & \quad\text{\scriptsize(Nonnegative Alloc.)}
\end{aligned}
\label{eqn:num_definition}\nonumber
\end{equation}
We highlight that NUM is \textit{\textbf{an ideal match}} to allocate ML resources in a multi-agent system because each agent has a utility, analogous to its accuracy, concave with its ML resources. By framing this as a NUM problem, we ensure a fair allocation among the agents derived from the Karush–Kuhn–Tucker (KKT) conditions. 

\textit{\underline{Related Work:}} 
Fairness-driven resource allocation mechanisms fall into three categories: \textit{leximin, max-min, and proportional fairness}. \textit{Leximin fairness} \cite{kurokawa2018leximin} minimizes disparities by sequentially allocating resources to agents with the highest demand, ensuring fairness but resulting in low total utility and quadratic complexity. \textit{Max-min fair} allocations \cite{ghodsi2011dominant} prioritize the least advantaged agents, but fail to capture diminishing marginal returns, prohibit resource substitutability \cite{varian2003intermediate}, and achieve low utility. \textit{Proportionally fair} allocations \cite{nagaraj2016numfabric} account for diminishing returns and provide better utility but often assume identical univariate logarithmic utility functions, limiting their applicability to heterogeneous settings. Unlike these three, \textit{our framework is \textbf{tailored for multi-agent ML} with heterogeneous resources}. 

\section{Problem Formulation}
Here, we propose a general formulation applicable to a wide range of collaborative scenarios involving a fleet of intelligent devices and a cloud, where constraints arise due to several factors, such as network limitations, power constraints, or labeling costs. This versatility makes our approach relevant for various applications, including autonomous fleets and ML-IoTs. We first describe the inference scenario with a univariate utility similar to the NUM. Then, we extend it to a multivariate utility for a training scenario to show its extensive use.

\subsection{Real-Time Inference in Cloud-Assisted Intelligent Fleet}\label{sec:infalloc}

We now formulate a practical scenario, shown in Fig. \ref{fig:toyex}, where a fleet of intelligent agents performs inference with their datasets. They utilize additional compute from the cloud to enhance the accuracy of their predictions. However, the cloud has limited resources. Our main objective is to determine the optimal resource allocation for each agent. Simply, we maximize the fleet's total accuracy by optimizing the allocated compute for inference with larger models.

\textbf{\underline{Univariate Utility Expression:}} Based on the result shown in Fig. \ref{fig:concvres}, we define the utility $U_i$ as agent $i$'s inference accuracy. The accuracy exhibits diminishing marginal returns as the cloud allocates more compute power (e.g., GPU FLOPS scaling the model size) $\rho_i$ in addition to the agent's local compute power $\rho_i^0$ for inference. This notion aligns with the Cobb-Douglas (Eq. \ref{eqn:multidimutility}) reformulated as:
\begin{equation}\small
    U_i(\rho_i;\rho_i^0,\gamma_i) = (\rho_i^0 + \rho_i)^{\gamma_i},\quad\forall i: \gamma_i \in (0,1).
    \label{eqn:EEN}
\end{equation}
 We propose that $\gamma_i$ is closely linked to the difficulty level (e.g., the noise) of the task performed by agent $i$. It captures task complexity. For $\gamma \approx 0$, simple tasks (e.g., CIFAR-10 in Fig. \ref{fig:concvres}-left) saturate quickly. For $\gamma \approx 1$, complex tasks (e.g., LLMs in Fig. \ref{fig:concvres}-right) scale linearly with resources. \textcolor{black}{We estimate $\gamma$ via fitting accuracy-resource curves inferred from edge metadata.}

\textbf{\underline{Problem 1:}} We maximize the total accuracy of the agents, $\sum_iU_i$, by allocating the cloud's compute power $\rho_i$ for agent $i$. Considering the cloud has limited compute power (GPU FLOPS) $P$ to serve for $N_r$ agents, we arrive at a formulation analogous to the NUM framework, as shown by:
\begin{equation}
\begin{aligned}
    \max_{\rho_1,...,\rho_{N_r}} & \sum_{i=1}^{N_r} U_i(\rho_i;\rho_i^0,\gamma_i),&\textrm{\scriptsize (Maximize Total Accuracy)}\\
     \text{s.t.} \quad& \sum_i^{N_r}{\rho_i} \leq P,&\text{\scriptsize (Compute Power Limit)}\\
    & \rho_i\geq0, \quad\forall i \in \{1,...N_r\}.&\text{\scriptsize (Nonnegative Compute Alloc.)}
\end{aligned}\label{eqn:EENopt}\nonumber
\end{equation} 
Problem 1 comprises additive concave functions in the objective and linear constraints, indicating a convex problem. We use a convex solver ({\tt CVXPY}) to solve it.

\textbf{\underline{Fairness and Its Physical Interpretation:}}
The optimal solution to this problem satisfies the KKT and Slater's conditions
. Using their mathematical implications, we derive a general fairness inequality 
as:
\begin{equation}
\begin{aligned}
    & \small -\nabla_\rho^\top U(\rho)(\rho - \rho^*) \geq 0,& \textrm{\scriptsize(Concave $U(\rho)$)}\\
    \end{aligned}\nonumber
\end{equation}
\begin{equation}
\begin{aligned}
    \Rightarrow \quad & \sum_{i=1}^{n} \gamma_i\frac{\rho_{i} - \rho_{i}^*}{(\rho_{i}^0+\rho_{i}^*)^{1-\gamma_i}} \leq 0. & \quad \textrm{\scriptsize(Proportional Fairness)}\nonumber
\end{aligned}
\end{equation}

This implies that when resources are fully utilized, allocating additional compute to one agent beyond its optimal share harms \textcolor{black}{the} total accuracy. The gain in proportional accuracy for that agent will be outweighed by the sum of the loss in proportional accuracy for the other. Prioritizing one agent at the expense of others reduces overall performance, embodying the concept of ``proportional fairness'' and indicates envy-freeness \cite{ghodsi2011dominant}. For the case of two-agent, agent $i$ and $j$, the fairness reduces to:
\begin{equation}\small
    \label{eqn:EEfairness1}
    \frac{\gamma_i}{(\rho_i^0 + \rho_i^*)^{1-\gamma_i}}=\frac{\gamma_j}{(\rho_j^0 + \rho_j^*)^{1-\gamma_j}}.
\end{equation}

Hence, the solution ensures Pareto optimality by lying on the Pareto frontier \cite{varian2003intermediate}, where no resource reallocation improves one agent's utility without reducing another's, denoted by $\frac{\partial U_i(\rho_i^*)}{\partial \rho_i} = \mu_0, \forall i;$ for Lagrange multiplier $\mu_0 > 0$. 

\begin{figure}[!t]
    \centering
    \includegraphics[width=0.7\linewidth]{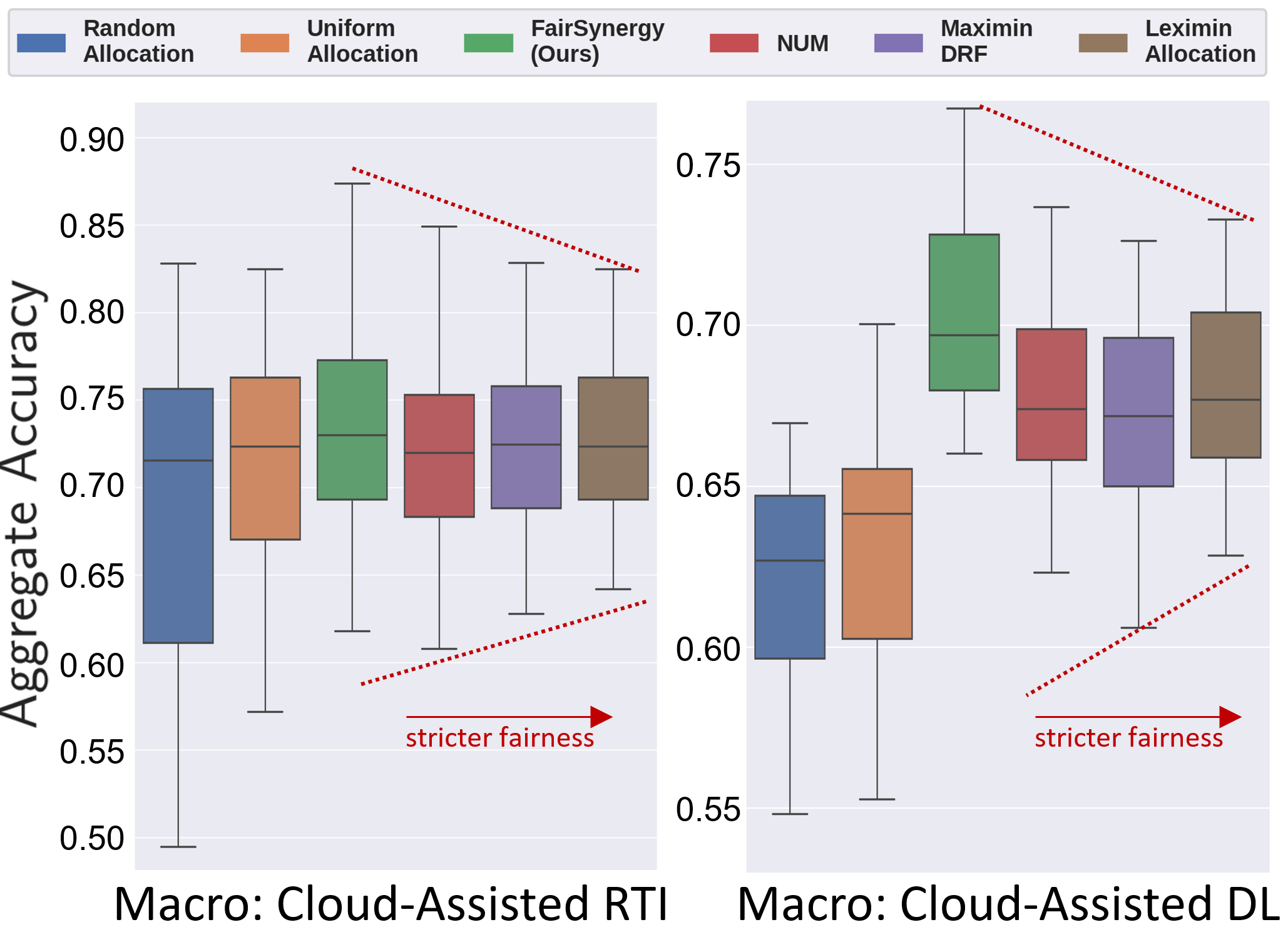}\caption{\small\textbf{How well does {\tt Fair-Synergy} perform compared to the benchmarks?} The simulations with {\tt CIFAR-10/100} datasets and {\tt MobileNet} models show {\tt Fair-Synergy} (green) consistently outperforms benchmarks. Specifically, in the RTI (left), {\tt Fair-Synergy} shows performance improvements of about 25\% in best-case and 6\% in worst-case scenarios. In the DL scenario (right), it exceeds benchmarks by roughly 11\% in best-case and 10\% in worst-case scenarios. Notably, stricter fairness methods (e.g., {\tt Leximin}) maximize \textcolor{black}{the} worst performance, while looser ones (e.g., {\tt NUM}, {\tt Fair-Synergy}) optimize the upper bound, highlighting the utility-fairness trade-off (red lines).
    \label{fig:boxplots}\vspace{-3mm}}
\end{figure}

\noindent \textit{\textbf{Decreasing inference effort for the agent with a larger local compute:}} When agents have equally hard queries $\gamma_{i,j}$, an increase in agent $i$'s local compute $\rho_i^0$(\textcolor{green}{$\uparrow$}) results in a decrease in \textcolor{black}{the} allocated compute for that agent $\rho_i$(\textcolor{black}{$\downarrow\downarrow$}), and an increase in allocated compute for the other agent $\rho_j$(\textcolor{black}{$\uparrow\uparrow$}), as: 
\begin{equation}
    \label{eqn:EEfairness2}\small
    \frac{\gamma_i}{(\rho_i^0 \textcolor{green}{\uparrow} + \rho_i^* \textcolor{black}{\downarrow\downarrow})^{1-\gamma_i}}=\frac{\gamma_j}{(\rho_j^0 + \rho_j^* \textcolor{black}{\uparrow\uparrow})^{1-\gamma_j}}.
\end{equation}

\noindent It ensures that the decision variables $\rho_i$ and $\rho_j$ are automatically updated by the optimization mechanism to maintain Eq. \ref{eqn:EEfairness2} under changing conditions, enforcing fair allocation.

\noindent \textit{\textbf{Increasing inference effort for the agent with harder queries:}} When the agents' local compute $\rho_{i,j}^0$ are equal, an increase in agent $i$'s hardness (e.g. harder-to-parse images or complex text prompts) $\gamma_i$(\textcolor{green}{$\uparrow$}) results in an increase in the cloud's allocated compute $\rho_i$(\textcolor{black}{$\uparrow\uparrow$}) for agent $i$, and a decrease in the allocated compute $\rho_j$(\textcolor{black}{$\downarrow\downarrow$}) for agent $j$, as:
\begin{equation}\small
    \label{eqn:EEfairness3}
    \frac{\gamma_i \textcolor{green}{\uparrow}}{(\rho_i^0 + \rho_i^*\textcolor{black}{\uparrow\uparrow})^{1-\gamma_i}}=\frac{\gamma_j}{(\rho_j^0 + \rho_j^*\textcolor{black}{\downarrow\downarrow})^{1-\gamma_j}}.
\end{equation}

\subsection{Distributed Learning in Cloud-Assisted Intelligent Fleet}\label{sec:trainalloc}
After the univariable, we now extend it to the multivariable.

\textbf{\underline{Multivariate Utility:}}
Drawing on \textcolor{black}{the} observations from Fig. \ref{fig:concvres}, we posit that the model accuracy $U$ is concave \textcolor{black}{in} an increasing number of annotated training samples $\alpha$ and model parameters scaling with the computing power $\rho$. We propose that \textit{\textbf{these decision variables are orthogonal}}. Hence, they form a utility similar to the Cobb-Douglas production function (e.g., the labeled data is capital and the compute power is labor) in Eq. \ref{eqn:multidimutility}. The training data for agent $i$ is the sum of the locally existing labeled data, $\alpha_i^0 \geq 0$, and the data annotated in the cloud $\alpha_i$ for agent $i$, either manually or with a massive high-confidence model. The total compute for agent $i$ is the sum of the agent's local compute power, $\rho_i^0 \geq 0$, and \textcolor{black}{the} allocated compute $\rho_i$ for the agent by the cloud. The utility for agent $i$, as a function of its resources, is formulated as:
\begin{equation}\small
U_i(\rho_i, \alpha_i; \rho_i^0, \alpha_i^0, \gamma_i^\rho, \gamma_i^\alpha) = (\rho_i^0 + \rho_i)^{\gamma_i^\rho}(\alpha_i^0 + \alpha_i)^{\gamma_i^\alpha}, \gamma_i^{\cdot} \in (0,1). \label{eqn:trainingutil}    
\end{equation}

\noindent $\gamma^\alpha$ and $\gamma^\rho$ represent \textcolor{black}{the} dataset and model complexity, respectively. For $\gamma^\alpha \approx 0$ (e.g., MNIST) and $\gamma^\rho \approx 0$ (e.g., EfficientNet), tasks show rapid performance saturation for increasing resources (Fig. \ref{fig:concvres}). For $\gamma^\alpha \approx 1$ (e.g., CIFAR-100) and $\gamma^\rho \approx 1$ (e.g., ResNet), the accuracy scales linearly.

\textbf{\underline{Problem 2:}} The algorithm aims to maximize the aggregate accuracy $\sum_i U_i$ of $N_r$ agents by optimally allocating cloud resources, including labeling labor $\alpha$ (limited by time $T$) and model size scaled by compute power $\rho$ (constrained by \textcolor{black}{the} total compute $P$). This problem is formulated as:
\begin{equation}
    \max_{ \substack{\rho_1, ., \rho_{N_r}\\\alpha_i, ., \alpha_{N_r}}}\quad\sum_{i=1}^{N_r} U_i(\rho_i, \alpha_i; \rho_i^0, \alpha_i^0, \gamma_i^\rho, \gamma_i^\alpha),\quad\quad\text{\scriptsize(Max. Total Acc.)}\nonumber    
\end{equation}
\begin{equation}
\begin{aligned}\small
    \text{s.t.}\quad &  \sum_{i=1}^{N_r}{\rho_i} \leq P,&\text{\scriptsize (Compute Power Limit)}\\ 
    &\sum_{i=1}^{N_r}{\alpha_i}\leq T,&\text{\scriptsize(Time Limit)}\\
    &\alpha_i \geq 0, \quad \forall i \in \{1,...,N_r\},& \text{\scriptsize(Nonnegative Labeling Alloc.)}\\
    &\rho_i \geq 0, \quad \forall i \in \{1,...,N_r\}.&\text{\scriptsize (Nonnegative Compute Alloc.)}
    \label{eqn:DLopt}\nonumber
\end{aligned}
\end{equation}
\textit{Biconcave functions are concave with respect to one variable when others are fixed}
The objective in Problem 2 is biconcave with linear constraints, forming a biconvex minimization problem solvable using the \textit{Alternate Convex Search} (ACS).

Univariate is multivariate's special case. Uniform data complexity reduces accuracy to a function of compute alone.

\textbf{\underline{Fairness and Its Physical Interpretation:}} Following the KKT
, we get the fairness for multi-agent DL, shown by:
\begin{equation}\small
    (\gamma_i^\alpha)\frac{(\rho_i^0 + \rho_i)^{\gamma_i^\rho} }{(\alpha_i^0 + \alpha_i)^{1- \gamma_i^\alpha} }= (\gamma_j^\alpha)\frac{(\rho_j^0 + \rho_j)^{\gamma_j^\rho}}{(\alpha_j^0 +  \alpha_j )^{1- \gamma_j^\alpha}}.
\end{equation}

This natural result of Problem 2 implies Pareto optimality.

\textit{\textbf{Decreasing labeling effort for the agent with a higher proportion of labeled data:}} When agents have identical compute (e.g., the same model size), the data annotation effort shifts toward the agent with a smaller proportion of labeled data. For example, if agent $i$ possesses a larger proportion of labeled data than agent $j$ ($\alpha_i^0$\textcolor{green}{$\uparrow$}), the cloud annotates less data for agent $i$ ($\alpha_i$\textcolor{blue}{$\downarrow\downarrow$}) and more data for agent $j$ ($\alpha_j$\textcolor{blue}{$\uparrow\uparrow$}). This fairness achieves higher total accuracy and minimizes the overfitting risk. Decision variables adapt to deviations in environmental variables, as denoted in:
\begin{equation} \small
    (\gamma_i^\alpha)\frac{(\rho_i^0 + \rho_i)^{\gamma_i^\rho} }{(\alpha_i^0 \textcolor{green}{\uparrow} + \alpha_i \textcolor{blue}{\downarrow\downarrow})^{1- \gamma_i^\alpha} }=
(\gamma_j^\alpha)\frac{(\rho_j^0 + \rho_j)^{\gamma_j^\rho}}{(\alpha_j^0 +  \alpha_j \textcolor{blue}{\uparrow\uparrow})^{1- \gamma_j^\alpha}}.
\end{equation}

\begin{figure}[!t]
    \centering
    \includegraphics[width=\linewidth]{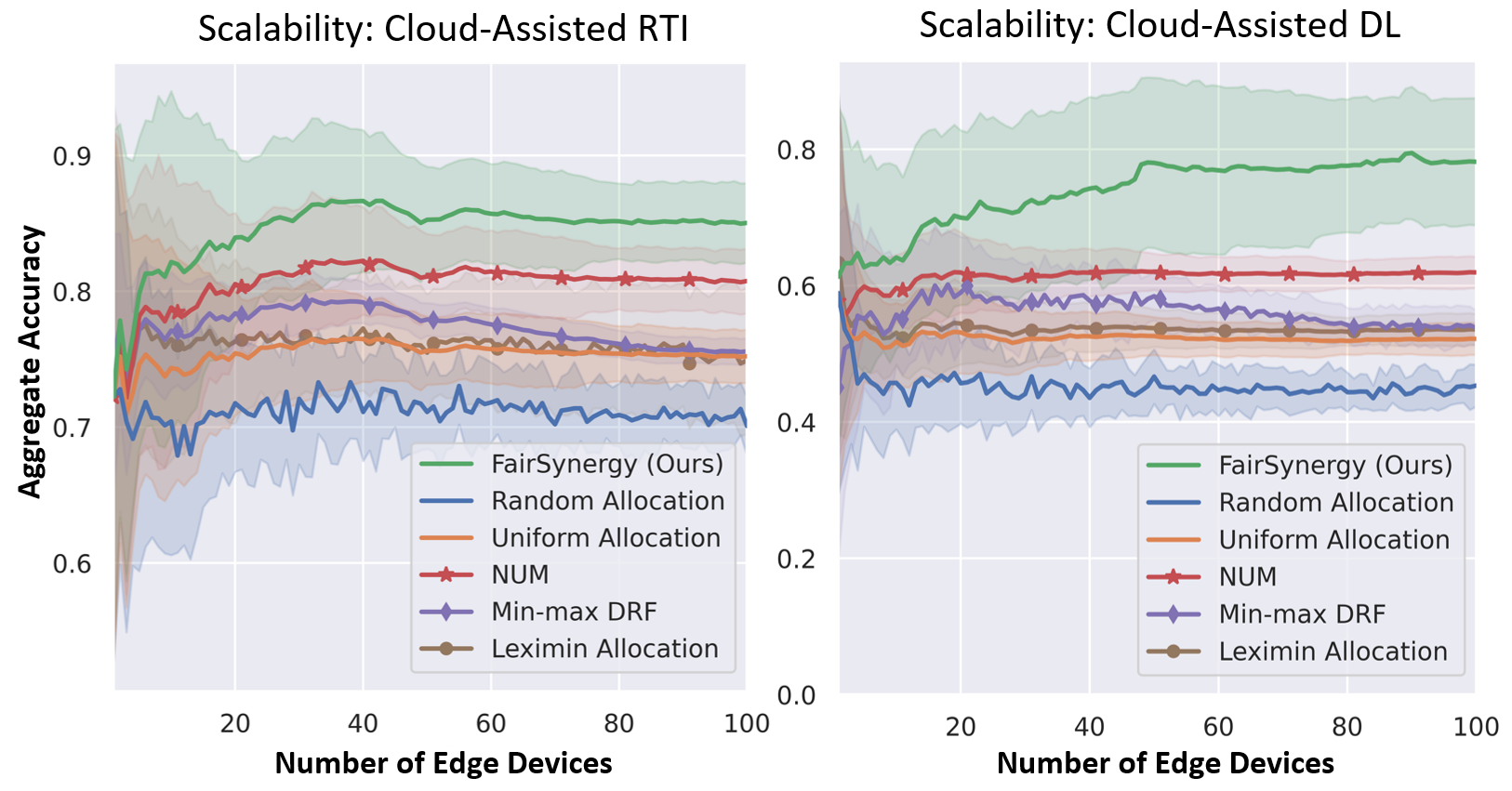}\caption{\small \textbf{How does {\tt Fair-Synergy} scale with increasing number of agents?} We study the effect of increasing edge devices with the models and datasets in Fig. \ref{fig:boxplots}. In RTI, it achieves progressively better performance, surpassing {\tt NUM} by 6\%, {\tt Uniform/DRF/Leximin} by 13\%, and {\tt Random} by 21\% with 100 agents. Gains plateau with more agents, with {\tt Uniform} outperforming {\tt Random}, and {\tt DRF} converging to uniform for increasing agents. In DL, {\tt Fair-Synergy} scales even better, outperforming {\tt NUM} by 30\%, {\tt Uniform/Leximin/DRF} by 51\%, and {\tt Random} by 76\%, showing superior scalability as the utility dimension increases. Relaxing fairness conditions results in higher overall utility due to the inherent performance-fairness tradeoff in both RTI and DL. \vspace{-3mm}\label{fig:scalability}} 
\end{figure}

\textit{\textbf{Increasing labeling effort for the agent with a larger compute power:}} Larger models have a greater capacity to utilize more training data. Thus, if agent $i$ has a larger model scaling up with a larger compute ($\rho_i \textcolor{green}{\uparrow}$) compared to agent $j$, the cloud annotates more data for agent $i$ ($\alpha_i \textcolor{blue}{\downarrow\downarrow}$) as the bigger model learns from more data without overfitting, as empirically shown in Fig. \ref{fig:concvres}. Conversely, agent $j$ with smaller compute receives less annotation ($\alpha_j \textcolor{blue}{\downarrow\downarrow}$) for efficient use of the cloud resources. Fairness imposes this by:
\begin{equation}\small
    (\gamma_i^\alpha)\frac{(\rho_i^0 + \rho_i)^{\gamma_i^\rho} \textcolor{green}{\uparrow}}{(\alpha_i^0+ \alpha_i \textcolor{blue}{\uparrow\uparrow})^{1- \gamma_i^\alpha} }=
(\gamma_j^\alpha)\frac{(\rho_j^0 + \rho_j)^{\gamma_j^\rho}}{(\alpha_j^0 +  \alpha_j \textcolor{blue}{\downarrow\downarrow})^{1- \gamma_j^\alpha}}.
\end{equation}
This implication of Problem 2 links the model size to its data consumption capacity for the ML resource allocation.

\section{Experiments}
In this section, we demonstrate how Fair-Synergy outperforms the baselines in both multi-agent inference and learning settings. The baseline methods assessed include:

{\tt Fair-Synergy} optimizes resource allocation based on the agents' local resources and task complexities to maximize total accuracy through the optimization problems in Eqs. \ref{eqn:EENopt} and \ref{eqn:DLopt} for the RTI and DL settings, respectively.

{\tt Random} allocation distributes resources randomly among \textcolor{black}{the} agents based on the cloud's maximum available compute and annotation labor. Specifically, $\sum_{i=1}^{N_r}\rho_i = P$ in the RTI setting, and $\sum_{i=1}^{N_r}\rho_i = P, \sum_{i=1}^{N_r}\alpha_i = T$ in the DL setting.

{\tt Uniform} allocation divides resources evenly by ignoring the agents' local properties. \textcolor{black}{The} agents receive $\rho_i = P/N_r$ in the RTI case and $\rho_i = P/N_r$, $\alpha_i = T/N_r$ in the DL case.

{\tt NUM} allots resources similarly to {\tt Fair-Synergy}, but has \textit{a uniform logarithmic objective} with equal $\gamma_i$ for all agents.

Dominant Resource Fair ({\tt DRF}) \cite{ghodsi2011dominant} equalizes dominant resource shares across agents and provides maximin fairness.

{\tt Leximin} allocation \cite{kurokawa2018leximin} imposes the \textit{strictest} resource fairness by iteratively prioritizing the least-advantaged agents.

\begin{figure}[t]
    \centering
    \includegraphics[width=1\linewidth]{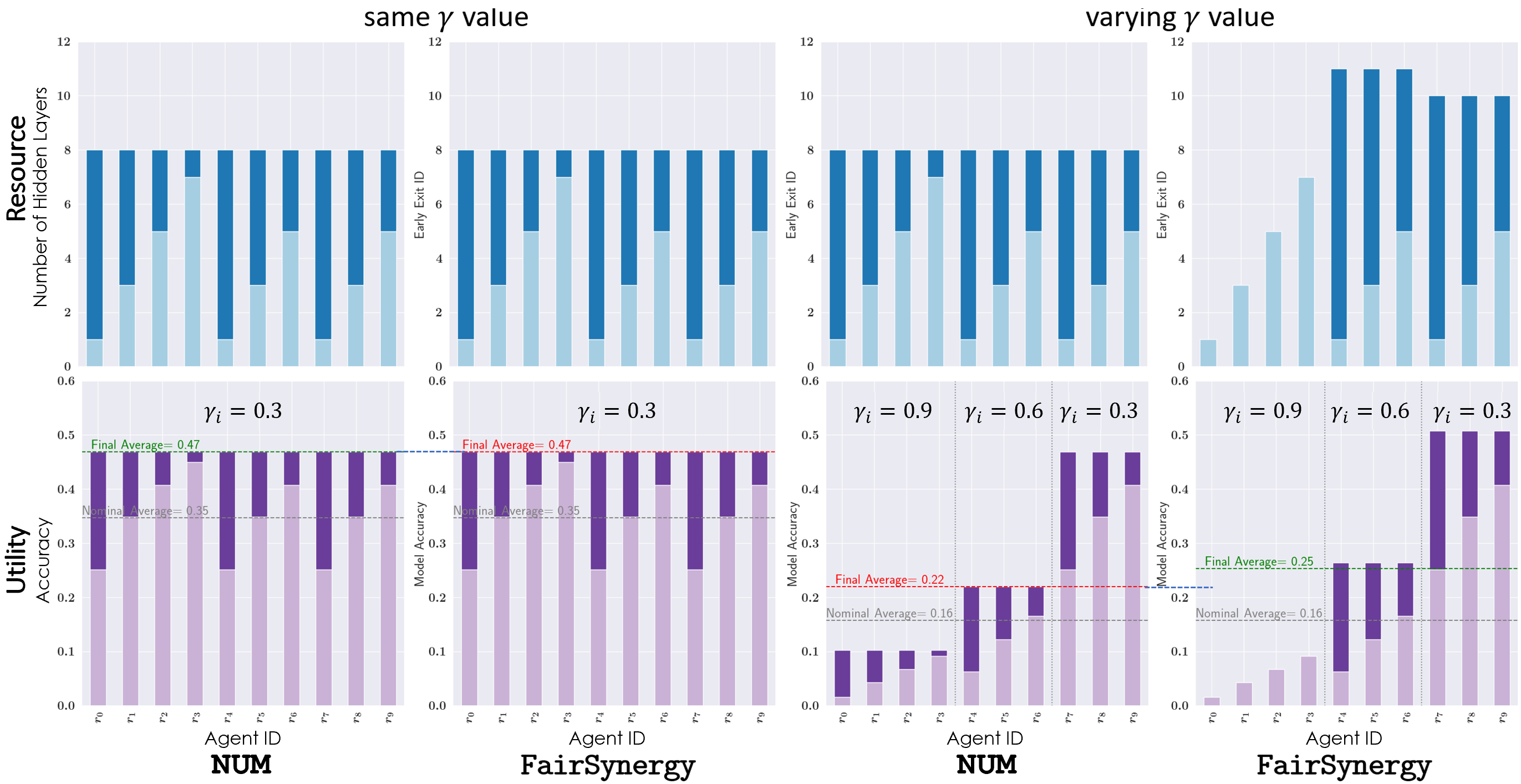}\caption{\small \textcolor{black}{\textbf{\textit{NUM vs. FairSynergy:}} In uniform hardness ($\gamma_i=0.3$; \textit{columns 1–2}), both methods allocate similarly and achieve comparable utility (classic water-filling). With heterogeneous hardness ($\gamma_i\in\{0.3,0.6,0.9\}$; \textit{columns 3–4}), NUM does not adapt its allocation, whereas ours reweights by agent elasticities and resources, adapting allocations for 14\% more utility in RTI. \vspace{-3mm}}}
    \label{fig:NUMCOMP}
\end{figure}

\noindent \textit{Evaluation Metric:} We evaluate the performance of {\tt Fair-Synergy} and the benchmarks based on both the \textbf{aggregate and individual accuracy} of the agents' ML models, measured pre- and post- resource allocation. We also assess scalability: the agent counts' effect on the performance.\\
\textit{\ul{How well does fair allocation adapt to heterogeneous resources of the multi-agent intelligence?}}\\
We randomly initialize (exogenous) system parameters and allocate resources across 1,000 experiments with varying seeds for statistical diversity. In each experiment, we evaluate the aggregate accuracies in two distinct settings: RTI and DL. For the RTI setting, the primary parameters influencing performance include the agents' local compute power $\rho_i^0$, the data complexities $\gamma_i$, and the cloud's total available compute resource $P$. Each agent's task-processing capability depends on these parameters, directly impacting resource allocation and overall system accuracy. In the DL setting, we expand the parameter set to include additional dimensions. Here, we consider the amount of labeled data $\alpha_i^0$ already available locally for each agent, in addition to \textcolor{black}{the} local compute power $\rho_i^0$, data complexity $\gamma_i$, the cloud's available labeling time $T$, and the total compute resource $P$. These variables reflect the DL environment complexity, where both compute and data annotations (time constraints) significantly impact training.\\
The results in Fig. \ref{fig:boxplots} demonstrate that {\tt Fair-Synergy} consistently outperforms the benchmarks, up to 25\% in RTI and 11\% in DL, because \textit{our approach dynamically adjusts its resource allocation in response to heterogeneous agent conditions, such as varying ML resources and data complexity}. The evaluation metric is concave. It means even small, linear improvements translate into significant enhancements\\
\textit{\ul{How well does fair resource allocation scale with increasing agent heterogeneity and additional resource dimensions?}}\\
We assess {\tt Fair-Synergy}'s scalability by increasing agents from 1 to 100 while proportionally scaling cloud resources, with $P = \rho \times N_r$ for the RTI setting and $P = \rho \times N_r$, $T = \alpha \times N_r$ for the DL setting. This maintains {\tt Uniform} allocation as a stable baseline for comparison. Each experiment is repeated 100 times with random seeds, and \textcolor{black}{the} results, including mean accuracy (solid lines) and standard deviation (shaded regions), are shown in Fig. \ref{fig:scalability}. Fig. \ref{fig:scalability} shows that {\tt Fair-Synergy} outperforms benchmarks by 21\% in RTI and 76\% in DL. Unlike other methods, \textit{its performance scales with more edges, as each adds heterogeneous features requiring adaptive allocation}. Its advantage is more pronounced in DL due to the inclusion of the additional resource dimension, labeled data, beyond compute power. \textcolor{black}{Specifically, NUM assumes uniform resource effects across agents. Our extended utility allows for heterogeneous impacts of different resources, and it is more adaptive, as in Fig. \ref{fig:NUMCOMP}}. This highlights {\tt Fair-Synergy}'s effectiveness in fleet ML.\\
\textcolor{black}{\textit{\underline{Complexity Analysis}}: For n agents and m resource types, Uniform and Random incur $O(mn)$. DRF runs in $O(nm+n\log n)$ to find the dominant from m shares across n agents and sorting those n. Leximin takes $O(n^2(m+\log n))$ as it solves n sequential max–min subproblems. NUM requires $O((mn)^3)$ for the interior point method and can be optimized to $O(mn\log n)$ by m water-filling with n sorting points. Ours converges in a few more iterations than NUM due to ACS.}

\section{Conclusion}
This paper introduces {\tt Fair-Synergy}, a fair resource allocation framework for intelligent agentic fleets. Our empirical analysis with the LLM and CV models ({\tt BERT}, {\tt VGG16}, {\tt MobileNet}, {\tt ResNets}) across the various datasets ({\tt MNIST}, {\tt CIFAR-10}, {\tt CIFAR-100}, {\tt BDD}, {\tt GLUE}) demonstrates a concave relationship between accuracy and agents' ML resources, such as model size, dataset size, and compute power. We incorporate the NUM to optimize the sum of concave accuracies, but extend beyond the limitations of NUM by employing a multivariate utility derived from the Cobb-Douglas form. This approach outperforms \textcolor{black}{the} benchmarks by up to $25\%$ in RTI and $11\%$ in DL with a small number of agents, and these margins increase more as the system scales for more agents. This underscores its efficiency in managing multiple resources and task complexities effectively, making it a robust resource allocation for multi-agent intelligence. Our code is publicly available.$^{\ref{coderef}}$ Future work will extend {\tt Fair-Synergy} to multi-round settings 
and explore a decentralized privacy-preserving dual form where agents bid for cloud resources.
\bibliographystyle{IEEEtran}
\bibliography{conference_101719}

\begin{thebibliography}{1}
\providecommand{\url}[1]{#1}
\csname url@samestyle\endcsname
\providecommand{\newblock}{\relax}
\providecommand{\bibinfo}[2]{#2}
\providecommand{\BIBentrySTDinterwordspacing}{\spaceskip=0pt\relax}
\providecommand{\BIBentryALTinterwordstretchfactor}{4}
\providecommand{\BIBentryALTinterwordspacing}{\spaceskip=\fontdimen2\font plus
\BIBentryALTinterwordstretchfactor\fontdimen3\font minus \fontdimen4\font\relax}
\providecommand{\BIBforeignlanguage}[2]{{%
\expandafter\ifx\csname l@#1\endcsname\relax
\typeout{** WARNING: IEEEtran.bst: No hyphenation pattern has been}%
\typeout{** loaded for the language `#1'. Using the pattern for}%
\typeout{** the default language instead.}%
\else
\language=\csname l@#1\endcsname
\fi
#2}}
\providecommand{\BIBdecl}{\relax}
\BIBdecl

\bibitem{kelly1998rate}
F.~P. Kelly, A.~K. Maulloo, and D.~K.~H. Tan, ``Rate control for communication networks: shadow prices, proportional fairness and stability,'' \emph{Journal of the Operational Research society}, vol.~49, pp. 237--252, 1998.

\bibitem{varian2003intermediate}
H.~R. Varian, \emph{Intermediate microeconomics: a modern approach}.\hskip 1em plus 0.5em minus 0.4em\relax Elsevier Brasil, 2003.

\bibitem{tan2019efficientnet}
M.~Tan and Q.~Le, ``Efficientnet: Rethinking model scaling for convolutional neural networks,'' in \emph{International conference on machine learning}.\hskip 1em plus 0.5em minus 0.4em\relax PMLR, 2019, pp. 6105--6114.

\bibitem{yu2020bdd100k}
F.~Yu \emph{et~al.}, ``Bdd100k: A diverse driving dataset for heterogeneous multitask learning,'' in \emph{Proceedings of the IEEE/CVF conference on computer vision and pattern recognition}, 2020, pp. 2636--2645.

\bibitem{nagaraj2016numfabric}
K.~Nagaraj, D.~Bharadia, H.~Mao, S.~Chinchali, M.~Alizadeh, and S.~Katti, ``Numfabric: Fast and flexible bandwidth allocation in datacenters,'' in \emph{2016 ACM SIGCOMM Conference}, 2016, pp. 188--201.

\bibitem{kurokawa2018leximin}
D.~Kurokawa, A.~D. Procaccia, and N.~Shah, ``Leximin allocations in the real world,'' \emph{ACM Transactions on Economics and Computation (TEAC)}, vol.~6, no. 3-4, pp. 1--24, 2018.

\bibitem{ghodsi2011dominant}
A.~Ghodsi \emph{et~al.}, ``Dominant resource fairness: Fair allocation of multiple resource types,'' in \emph{8th USENIX symposium on networked systems design and implementation (NSDI 11)}, 2011.

\end{thebibliography}

\end{document}